\documentclass[10pt, a4paper]{article}
\pdfoutput=1
\usepackage{lrec2020}
\usepackage{multibib}
\newcites{languageresource}{~}
\bibliographystylelanguageresource{lrec2020}
\usepackage{graphicx}
\usepackage{tabularx}
\usepackage{soul}
\usepackage{changepage}
\usepackage{enumitem}
\usepackage{xstring}

\usepackage[utf8]{inputenc}
\usepackage[LFE,LAE,T1]{fontenc} 
\usepackage{arabtex}
\usepackage{utf8}
\setcode{utf8}
\usepackage{tipa} 

\title{\emph{TArC}: Incrementally and Semi-Automatically Collecting a Tunisian Arabish Corpus}

\name{Elisa Gugliotta$^{1,2}$, Marco Dinarelli$^{1}$}

\address{1. LIG, B\^atiment IMAG - 700 avenue Centrale - Domaine Universitaire de Saint-Martin-d’Hères, France\\
2. University of Rome \emph{Sapienza}, Piazzale Aldo Moro 5, 00185 Roma, Italy \\
         elisa.gugliotta@uniroma1.it, marco.dinarelli@univ-grenoble-alpes.fr\\}

\abstract{
This article describes the constitution process of the first morpho-syntactically annotated Tunisian \emph{Arabish} Corpus (TArC). 
Arabish, also known as \emph{Arabizi}, is a spontaneous coding of Arabic dialects in Latin characters and \textit{arithmographs} (numbers used as letters). This \emph{code-system} was developed by Arabic-speaking users of social media in order to facilitate the writing in the Computer-Mediated Communication (CMC) and text messaging informal frameworks. 
There is variety in the realization of Arabish amongst dialects, and each Arabish code-system is under-resourced, in the same way as most of the Arabic dialects. In the last few years, the 
focus on Arabic dialects in the NLP field has considerably increased. Taking this into consideration, TArC will be a useful support for different types of analyses, computational and linguistic, as well as for NLP tools training. 
In this article we will describe preliminary work on the TArC semi-automatic construction process and some of the first analyses we developed on TArC. In addition, in order to provide a complete overview of the challenges faced during the building process, we will present the main Tunisian dialect characteristics and their encoding in Tunisian Arabish.
\newline  }

\begin{document}
\maketitleabstract

\section{Introduction}
\label{sec:intro}
Arabish is the romanization of Arabic Dialects (ADs) used for informal messaging, especially in social networks.\footnote{Also known as Arabizi (from the combination between the Arabic words \textipa{"}Arab\textipa{"}, \textipa{["Qarab]} and \textipa{"}English\textipa{"} \textipa{[i:n"Zli:zi:]}, or the English one: \textipa{"}easy\textipa{"}, \textipa{["i:zi]}), Franco-Arabic, Arabic Chat Alphabet, ACII-ized Arabic, and many others. \textipa{"}Arabish\textipa{"} is probably the result of the union between  \textipa{["Qarab]} and \textipa{"}English\textipa{"}.} This writing system provides an interesting ground for linguistic research, computational as well as sociolinguistic, mainly due to the fact that it is a spontaneous representation of the ADs, and because it is a linguistic phenomenon in constant expansion on the web. 
Despite such potential, little research has been dedicated to Tunisian Arabish (TA). In this paper we describe the work we carried to develop a flexible and multi-purpose TA resource. This will include a TA corpus, together with some tools that could be useful for analyzing the corpus and for its extension with new data.

First of all, the resource will be useful to give an overview of the TA. At the same time, it will be a reliable representation of the Tunisian dialect (TUN) evolution over the last ten years: the collected texts date from 2009 to present. This selection was done with the purpose to observe to what extent the TA orthographic system has evolved toward a writing convention.
Therefore, the TArC will be suitable for phonological, morphological, syntactic and semantic studies, both in the linguistic and the Natural Language Processing (NLP) domains.
For these reasons, we decided to build a corpus which could highlight the structural characteristics of TA through different annotation levels, including Part of Speech (POS) tags and lemmatization. In particular, to facilitate the match with the already existing tools and studies for the Arabic language processing, we provide a transcription in Arabic characters at token level, following the Conventional Orthography for Dialectal Arabic guidelines 
\emph{CODA*} (CODA \emph{star}) \cite{habash-etal-2018-unified} and taking into account the specific guidelines for TUN (CODA TUN) \cite{DBLP:conf/lrec/ZribiBMEBH14}. 
Furthermore, even if the translation is not the main goal of this research, we have decided to provide an Italian translation of the TArC’s texts.\footnote{We considered the Italian translation as an integrated part of the annotation phase that would have cost us less effort in addition to us if carried out in our mother-tongue. The possibility of a TArC English translation is left open for a later time.} 

Even though in the last few years ADs have received an increasing attention by the NLP community, many aspects have not been studied yet and one of these is the Arabish code-system. The first reason for this lack of research is the relatively recent widespread of its use: before the advent of the social media, Arabish usage was basically confined to text messaging.
However, the landscape has changed considerably, and particularly thanks to the massive registration of users on Facebook since 2008. At that time, in Tunisia there were still no Arabic keyboards, neither for Personal Computers, nor for phones, so Arabic-speaking users designed TA for writing in social media (Table \ref{tab1}).  
A second issue that has held back the study of Arabish is its lack of a standard orthography, and the informal context of use. It is important to note that also the ADs lack a standard code-system, mainly because of their oral nature.
In recent years the scientific community has been active in producing various sets of guidelines for dialectal Arabic writing in Arabic characters: CODA (Conventional Orthography for Dialectal Arabic)  \cite{habash-etal-2012-conventional}.
\newline

The remainder of the paper is organized as follows:  section~\ref{sec:StART} is an overview of NLP studies on TUN and TA; section~\ref{sec:TD_TA} describes TUN and TA; section~\ref{sec:TArC} presents the TArC corpus building process; section~\ref{sec:pro} explains preliminary experiments with a semi-automatic transcription and annotation procedure, adopted for a faster and simpler construction of the TArC corpus; conclusions are drawn in section~\ref{sec:concl}

\section{Related Work}
\label{sec:StART}

In this section, we provide an overview of work done on automatic processing of TUN and TA. As briefly outlined above, many studies on TUN and TA aim at solving the lack of standard orthography. The first Conventional Orthography for Dialectal Arabic (CODA) was for Egyptian Arabic  \cite{habash-etal-2012-conventional} and it was used by 
 \newcite{bies2014transliteration} for Egyptian Arabish transliteration into Arabic script.
The CODA version for TUN (CODA TUN) was developed by  \newcite{DBLP:conf/lrec/ZribiBMEBH14}, and was used in many studies, like \newcite{boujelbane2015traitements}. Such work presents a research on automatic word recognition in TUN. Narrowing down to the specific field of TA, CODA TUN was used in \newcite{masmoudi2015arabic} to realize a TA-Arabic script conversion tool, implemented with a rule-based approach.
The most extensive CODA is CODA*, a unified set of guidelines for 28 Arab city dialects  \cite{habash-etal-2018-unified}. For the present research,  CODA* is considered the most convenient guideline to follow due to its extensive applicability, which will support comparative studies of corpora in different ADs. 
As we already mentioned, there are few NLP tools available for Arabish processing in comparison to the amount of NLP tools realized for Arabic.
Considering the lack of spelling conventions for Arabish, previous effort has focused on automatic transliteration from Arabish to Arabic script, e.g.  \newcite{chalabi2012romanized}, \newcite{darwish2013arabizi}, and \newcite{al2014automatic}. 
These three work are based on a character-to-character mapping model that aims at generating a range of alternative words that must then be selected through a linguistic model. A different method is presented in  \newcite{younes2018sequence}, in which the authors present a sequence-to-sequence-based approach for TA-Arabic characters transliteration in both directions \cite{Sutskever-2014-SSL-2969033.2969173,younes2018sequence}.

Regardless of the great number of work done on TUN automatic processing, there are not a lot of TUN corpora available for free  \cite{younes2018survey}.
To the best of our knowledge there are only five TUN corpora freely downloadable: one of these is the PADIC  \citelanguageresource{PADIC}, composed of 6,400 sentences in six Arabic dialects, translated in \emph{Modern Standard Arabic} (MSA), and annotated at sentence level.\footnote{The Arabic dialects of the PADIC are: TUN (Sfax), two dialects of Algeria, Syrian, Palestinian and Moroccan \cite{meftouh2018padic}.} Two other corpora are the Tunisian Dialect Corpus Interlocutor (TuDiCoI) \citelanguageresource{Tudicoi} and the Spoken Tunisian Arabic Corpus (STAC) \citelanguageresource{stac}, which are both morpho-syntactically annotated. The first one is a spoken task-oriented dialogue corpus, which gathers a set of conversations between staff and clients recorded in a railway station. TuDiCoI consists of 21,682 words in client turns \cite{graja2013discriminative}.\footnote{The annotation was carried out only for 7,814 word.}
The STAC is composed of 42,388 words collected from audio files downloaded from the web (as TV channels and radio stations files)  \cite{zribi2015spoken}. A different corpus is the TARIC \citelanguageresource{Taric}, which contains 20 hours of TUN speech, transcribed in Arabic characters \cite{masmoudi2014corpus}.\footnote{The 20 hours recorded are equivalent to 71,684 words.}
 The last one is the TSAC \citelanguageresource{Tsac}, containing 17k comments from Facebook, manually annotated to positive and negative polarities \cite{medhaffar2017sentiment}. This corpus is the only one that contains TA texts as well as texts in Arabic characters. As far as we know there are no available corpora of TA transcribed in Arabic characters which are also morpho-syntactically annotated.
In order to provide an answer to the lack of resources for TA, we decided to create TArC, a corpus entirely dedicated to the TA writing system, transcribed in CODA TUN and provided with a lemmatization level and POS tag annotation. 

\section{Characteristics of Tunisian Arabic and Tunisian Arabish}
\label{sec:TD_TA}

The Tunisian dialect (TUN) is the spoken language of Tunisian everyday life, commonly referred to as \RL{الدَّارِجَة}, \textit{ad-dārija}, \RL{العَامِّيَّة}, \textit{al-‘āmmiyya}, or \RL{التُّونْسِي}, \textit{\textipa{@t-tūnsī}}.
According to the traditional diatopic classification, TUN belongs to the area of Maghrebi Arabic, of which the other main varieties are Libyan, Algerian, Moroccan and the 
\textsubdot{H}assānīya variety of Mauritania\footnote{The main geographical macro-areas, also called geolects, are the area of the Levant or Syro-Palestinian, Egypt and Sudan, Mesopotamia, Maghreb (North Africa) and the Arabian Peninsula. 
}  \cite{durand2009dialettologia}.
Arabish is the transposition of ADs, which are mainly spoken systems, into written form, thus turning into a quasi-oral system (this topic will be discussed in section  \ref{subsec:TA_system}). In addition, Arabish is not realized through Arabic script and consequently it is not subject to the Standard Arabic orthographic rules. As a result, it is possible to consider TA as a faithful written representation of the spoken TUN \cite{akbar2019arabizi}. 

\subsection{Tunisian Arabic}

The following list provides an excerpt of the principal features of TUN, which, through the TArC, would be researched in depth among many others. \footnote{For a detailed description please refer to \cite{durand2009dialettologia}, \cite{marccais1977esquisse}.}

At the phonetic level, some of the main characteristics of TUN, and Maghrebi Arabic in general, are the following: \\
\begin{adjustwidth}{1em}{0pt}
\textbf{*} Strong influence of the Berber substratum, to which it is possible to attribute the conservative phonology of TUN consonants.
\end{adjustwidth}

\begin{adjustwidth}{1em}{0pt}
\textbf{*} Presence of new emphatic phonemes, above all [\textsubdot{r}], [\textsubdot{l}], [\textsubdot{b}].\\  
\textbf{*} Realization of the voiced post-alveolar affricate [\textdyoghlig]  as fricative \textipa{[Z]}. \\
\textbf{*} Overlapping of the pharyngealized voiced alveolar stop \textipa{[d\super Q]}, <\RL{ض}>, with the fricative \textipa{[D\super Q]}, <\RL{ظ}>. \\
\textbf{*} Preservation of a full glottal stop \textipa{[P]} mainly in cases of loans from Classical Arabic (CA) or exclamations and interjections of frequent use.
\textbf{*} Loss of short vowels in open syllables.
\newpage
\textbf{*} Monophthongization.\footnote{Reduction of the diphthongs \textipa{[aw]} and \textipa{[aj]} to \textipa{[u:]} and \textipa{[i:]} in pre-Hilalian dialects, and to \textipa{[o:]} and \textipa{[e:]} in the Hilalian ones.} In TUN <\RL{بَيت}>, \textipa{["baijt]}, \textipa{"}house\textipa{"}, becomes \textipa{["bi:t]} meaning \textipa{"}room\textipa{"}. \\ 
\textbf{*} Palatalization of ā: Imāla, <\RL{إمالة}>, literally \textipa{"}inclination\textipa{"}.
(In TUN the phenomenon is of medium intensity.)
Thereby the word <\RL{باب}>, \textipa{["ba:b]}, \textipa{"}door\textipa{"},
becomes \textipa{["bE:b]}. \\ 
\textbf{*} Metathesis. 
(Transposition of the first vowel of the word.
It occurs when non-conjugated verbs or names without suffix begin with the
sequence CCvC, where C stands for ungeminated consonant, and
\textipa{"}v\textipa{"} for short vowel. When a suffix is added to this type of
name, or a verb of this type is conjugated, the first vowel changes position
giving rise to the CvCC sequence.)
In TUN it results in: \textipa{"}(he) has
understood\textipa{"}: \\ <\RL{فْهِم}>, \textipa{["fh@m]}, \textipa{"}(she) has
understood\textipa{"}: <\RL{فِهْمِت}>, \textipa{["f@hm@t]} or
\textipa{"}leg\textipa{"}: <\RL{رْجِل}>, \textipa{["rZ@l]}, \textipa{"}my
leg\textipa{"}: <\RL{رِجْلِي}>, \textipa{["r@Zli]}.\\ 

\end{adjustwidth} 

Regarding the morpho-syntactic level, TUN presents:\\
\begin{adjustwidth}{1em}{0pt}
\textbf{*} Addition of the prefix /-n/ to first person verbal morphology in \textit{mu\textsubdot{d}āri'} (imperfective).\\
\textbf{*} Realization of passive-reflexive verbs through the morpheme /-t/ \footnote{The morpheme /-t/ can be traced back to the same morpheme present in the V and VI verbal patterns of CA \cite{mion2004osservazioni}.} prefixed to the verb as in the example:\\ <\RL{سوريّة مالحَفْصيّة تْتِلْبِس}>, 
\textipa{[su:"ri:j:a m@l-\textcrh af"s\super Qij:a t-"t@lb@s]}, \textipa{"}the shirts of \textsubdot{H}af\textsubdot{s}iya\footnote{ \textsubdot{H}af\textsubdot{s}iya is a neighborhood in the Medīna of Tunis, known for its great daily frīp (second-hand market).} are not bad\textipa{"}, (lit: \textipa{"}they dress\textipa{"}). \\
\textbf{*} Loss of gender distinction at the 2\super{nd} and 3\super{rd} persons, at verbal and pronominal level. \\
\textbf{*} Disappearance of the dual form from verbal and pronominal inflexion.
There is a residual of pseudo-dual in some words fixed in time in their dual form. \\
\textbf{*} Loss of relative pronouns flexion and replacement with the invariable form <\RL{اِلّي}>, \textipa{[@l:i]}.\\
\textbf{*} Use of presentatives  /\textsubdot{r}ā-/ and /hā-/ with the meaning of \textipa{"}here\textipa{"}, \textipa{"}look\textipa{"}, as in the example in TUN: <\RL{راني مَخْنوق}>,
\textipa{["}\textsubdot{r}\textipa{a:ni: m@x"nu:q]}, \textipa{"}here I am asphyxiated (by problems)\textipa{"}, or in <\RL{هاك دَبَّرْتْها}>, \textipa{["ha:-k d@"b:@rt-ha:]}, \textipa{"}here you are, finding it (the solution)\textipa{"} hence: \textipa{"}you were lucky\textipa{"}. \\
\textbf{*} Presence of circumfix negation marks, such as <~<\RL{ما}>, \textipa{[ma]} + verb + <\RL{ش}>, \textipa{[S]}>. The last element of this structure must be omitted if there is another negation, such as the Tunisian adverb <\RL{عُمْر}>, \textipa{["Qomr]}, \textipa{"}never\textipa{"}, as in the structure: <\textipa{["Qomr]} + personal pronoun suffix + \textipa{[m@]} + perfect verb>. This construction is used to express the concept of \textipa{"}never having done\textipa{"} the action in question, as in the example:
<\RL{عُمري ما كُنْت  نِتْصَوُّر...}>, \textipa{["Qomr-i ma "k@nt n@ts\super Qaw:@r]}, \textipa{"}I never imagined that...\textipa{"}. 
\newline Instead, to deny an action pointing out that it will never repeat itself again, a structure widely used is <[ma] + \textipa{["Qa:d]} + \textipa{[S]} + imperfective verb>, where the element within the circumfix marks is a grammaticalized  element of verbal origin from CA: <\RL{عاد}>, \textipa {["Qa:d]}, meaning \textipa{"}to go back, to reoccur\textipa{"}, which gives the structure a sense of denied repetitiveness, as in the sentence:
\newline <\RL{هو ما عادِش يَرْجَع}>,   \textipa{["hu:wa ma "Qa:d-S "j@rZaQ]}, 

\textipa{"}he will not come back\textipa{"}. 
\newline Finally, to deny the nominal phrase, in TUN both the
<\RL{موش}>, \textipa{["mu:S]}, and the circumfix marks are frequently used. 
For the negative form of the verb \textipa{"}to be\textipa{"} in the present, circumfix marks can be combined with the personal suffix pronoun, placed between the marks, as in <\RL{مَانِيش}>, \textipa{[ma"ni:S]}, \textipa{"}I am not\textipa{"}.\\ Within the negation marks we can also find other types of nominal structures, such as: <\textipa{[fi:]} + \textipa{["bE:l]}(\textipa{"}mind\textipa{"})
+ personal pronoun suffix>, which has a value equivalent to the verb \textipa{"}be aware of\textipa{"}, as in the example: \\ <\RL{ما في باليش}>, \textipa{[ma fi: bE:l-"i:-S]}, \textipa{"}I did not know\textipa{"}.
\end{adjustwidth}

\subsection{Tunisian Arabish}
\label{subsec:TA_system}

As previously mentioned, we consider Arabish a quasi-oral system.
With \textit{quasi-orality} it is intended the form of communication typical of Computer-Mediated Communication (CMC), characterized by informal tones, dependence on context, lack of attention to spelling and especially the ability to create a sense of collectivity \cite{hert1999quasi}\footnote{Even though the CMC is generally a type of asynchronous communication.}. 

TA and TUN have not a standard orthography, with the exception of the CODA TUN. Nevertheless, TA is a spontaneous code-system used since more than ten years, and is being conventionalized by its daily usage. 

From the table~\ref{tab1}, where the coding scheme of TA is illustrated, it is possible to observe that there is no one-to-one correspondence between TA and TUN characters and that often Arabish presents overlaps in the encoding possibilities. 
The main issue is represented by the not proper representation by TA of the emphatic phones: \textipa{{[D\super Q]}}, \textipa{[t\super Q]} and \textipa{[s\super Q]}.

\begin{table}[htbp]
\begin{center}
\begin{tabularx}{\columnwidth}{c|c|c||c|c|c}

      \hline
      \textit{IPA}& \textit{TUN} &  \textit{TA} & \textit{IPA}& \textit{TUN}& \textit{TA} \\
      \hline
      \hline
      &&&&& \\
      \textipa{[a:]} & \RL{ة} & a, e, h & \textipa{[a][a:]} & \RL{ى}, \RL{ا} & a, e, é, è\\
     
      \textipa{[P]} & \RL{ء} & 2 & \textipa{[D\super Q]} & \RL{ض} & dh,  th, d\\ 
     
      \textipa{[b]} & \RL{ب} & b, p & \textipa{[t\super Q]} & \RL{ط} & 6, t\\
      
      \textipa{[t]} & \RL{ت} & t & \textipa{[D\super Q]} & \RL{ظ} & th, dh\\
      
      \textipa{[T]} & \RL{ث} & th & \textipa{[Q]} & \RL{ع} & 3, a\\
      
      \textipa{[Z]} & \RL{ج} & j & \textipa{[G]} & \RL{غ} & 4, gh\\
      
      \textipa{[\textcrh]} & \RL{ح} & 7, h & \textipa{[f]} & \RL{ف} & f\\
      
      \textipa{[x]} & \RL{خ} & 5, kh & \textipa{[q]} & \RL{ق} & 9, q\\
      
      \textipa{[d]} & \RL{د} & d & \textipa{[k]} & \RL{ك} & k\\
      
      \textipa{[D]} & \RL{ذ} & dh & \textipa{[l]} & \RL{ل} & l\\
      
      \textipa{[r]} & \RL{ر} & r & \textipa{[m]} & \RL{م} & m\\
      
      \textipa{[z]} & \RL{ز} & z & \textipa{[n]} & \RL{ن} & n\\
      
      \textipa{[s]} & \RL{س} & s & \textipa{[h]} & \RL{ه} & 8, h\\
      
      \textipa{[S]} & \RL{ش} & ch, (sh) & \textipa{[w][u:]} & \RL{و} & ou, w\\
      
      \textipa{[s\super Q]} & \RL{ص} &  s &  \textipa{[j][i:]} & \RL{ي} & i, y\\
      &&&&& \\
      \hline

\end{tabularx}
\caption{Arabish code-system for TUN}
\label{tab1}
 \end{center}
\end{table}

On the other hand, being TA not codified through the Arabic alphabet, it can well represent the phonetic realization of TUN, as shown by the
following examples: 

\textbf{ *} The Arabic alphabet is generally used for formal
conversations in Modern Standard Arabic (MSA), the Arabic of formal situations, or in that of 
Classical Arabic (CA), the Arabic of the Holy Qur’ān, also known as
‘The Beautiful Language’. Like MSA and CA, also Arabic Dialects
(ADs) can be written in the Arabic alphabet, but in this case it is
possible to observe a kind of hypercorrection operated by the speakers in order to respect the writing rules of MSA. For example, in TUN texts written in Arabic script, it is possible to find a ‘silent vowel’ (namely an epenthetic \textipa{"}alif
<\RL{ا}>) written at the beginning of those words starting with
the sequence ‘\#CCv’, which is not allowed in MSA. 

\textbf{ *} Writing TUN in Arabic script, the Code-Mixing or Switching in foreign language will be unnaturally reduced. 

\textbf{ *} As described in table~\ref{tab1}, the Arabic alphabet
is provided with three short vowels, which correspond to the three long ones: \textipa{[a:]}, \textipa{[u:]}, \textipa{[i:]}, but TUN presents
a wider range of vowels. Indeed, regarding the early presented
characteristics of TUN, the TA range of vowels offers better possibility to represent most  of the TUN characteristics outlined in the previous subsection, in particular:

\begin{itemize}[nosep]
    \item Palatalization.
    \item Vowel metathesis.
    \item Monophthongization.\footnote{Regarding the last two phenomena, they can be visible in Arabic script only in case of texts provided with short vowels, which are quite rare.}
\end{itemize}

\section{Tunisian Arabish Corpus}
\label{sec:TArC}

In order to analyze the TA system, we have built a TA Corpus based on social media data, considering this as the best choice to observe the quasi-oral nature of the TA system.

\subsection{Text collection}
\label{subsec:txt-coll}
The corpus collection procedure is composed of the following steps: 
\begin{enumerate}[nosep]
\item Thematic categories detection.
\label{step1} 
\item Match of categories with sets of semantically related TA keywords.
\label{step2}
\item Texts and metadata extraction. \\
\label{step3}
\end{enumerate}

\textbf{Step~\ref{step1}.} In order to build a Corpus that was as representative as possible of the linguistic system, it was considered useful to identify wide thematic categories that could represent the most common topics of daily conversations on CMC.

In this regard, two instruments with a similar thematic organization have been employed: 
\begin{itemize}[nosep]
\item\textbf{‘A Frequency Dictionary of Arabic’} \\ \cite{buckwalter2014frequency} In particular its ‘Thematic Vocabulary List’ (TVL). 
\item\textbf{‘Loanword Typology Meaning List’} \\ A list of 1460 meanings\footnote{The ‘Loanword Typology Meaning List’ is a result of a joint project by Uri Tadmor and Martin Haspelmath: the ‘Loanword Typology Project’ (LWT), launched in 2004 and ended in 2008.} (LTML)  \cite{haspelmath2009loanwords}. 
\end{itemize}
The TVL consists of 30 groups of frequent words, each one represented by a thematic word.
The second consists of 23 groups of basic meanings sorted by representative word heading.
Considering that the boundaries between some categories are very blurred, some categories have been merged, such as \textipa{"}Body\textipa{"} and \textipa{"}Health\textipa{"}, (see table~\ref{tab2}). Some others have been eliminated, being not relevant for the purposes of our research, e.g. \textipa{"}Colors\textipa{"}, \textipa{"}Opposites\textipa{"}, \textipa{"}Male names\textipa{"}. In the end, we obtained 15 macro-categories listed in table~\ref{tab2}. \\

\textbf{Step~\ref{step2}.} Aiming at easily detect texts and the respective \textit{seed URLs}, without introducing relevant query biases, we decided to avoid using the category names as query keywords \cite{schafer2013web}. Therefore, we associated to each category a set of TA keywords belonging to the basic Tunisian vocabulary. We found that 
a semantic category with three meanings was enough to obtain a sufficient number of keywords and URLs for each category. For example, to the category \textipa{"}Family\textipa{"} the meanings: \textipa{"}son\textipa{"}, \textipa{"}wedding\textipa{"}, \textipa{"}divorce\textipa{"} have been associated in all their TA variants, obtaining a set of 11 keywords (table~\ref{tab2}).\\ 

\begin{table}[htbp]
\begin{center}
\begin{tabularx}{\columnwidth}{|X|X|} 

      \hline
      \textbf{Macro-Categories}& \textbf{Words Associated} \\
      \hline
      1. Family \newline\textit{son, wedding, divorce}& weld, wild, 3ars, 3ers, \newline tla9, 6la9, tlaq, 6laq, tle9, tleq, 6leq\\
      \hline
      2. Clothing \newline\textit{dress, shoes, t-shirt} &robe, lebsa, rouba, \newline sabat, spedri, spadri, \newline marioul, maryoul, \newline meryoul, merioul\\
      \hline
      3. Automobiles \newline\textit{gasoil, engine, \newline occasion} & mazout, motor, moteur, motour, forsa\\
      \hline
      4. Animals \newline\textit{cock, dog, cat} & sardouk, kelb, kalb, \newline9attous, gattous\\
      \hline
      5. Body and Health \newline\textit{sick, doctor, health} & maridh, marith, mridh, ettbib, tbib, sa77a, sa7a, sahha, saha  \\
      \hline

\end{tabularx}
\caption{Example of the fifteen thematic categories}
\label{tab2}
 \end{center}
\end{table}

\textbf{Step~\ref{step3}.} 
We collected about 25,000 words and the related metadata as first part of our corpus, which are being semi-automatically transcribed into Arabic characters (see next sections).
We planned to increase the size of the corpus at a later time.
Regarding the metadata, we have extracted the information published by users, focusing on the three types of information generally used in ethnographic studies: 
\begin{enumerate}[nosep]
\item Gender: Male (M) and Female (F). 
\item Age range: [10-25], [25-35], [35-50], [50-90]. 
\item City of origin.
\end{enumerate}

\subsection{Corpus Creation}
\label{subsec:Corpus_cr}

In order to create our corpus, we applied a word-level annotation. This phase was preceded by some data pre-processing steps, in particular tokenization. 
Each token has been associated with its annotations and metadata (table~\ref{tab3}). 
In order to obtain the correspondence between Arabish and Arabic morpheme transcriptions, tokens were segmented into morphemes.
This segmentation was carried out completely manually for a first group of tokens.\footnote{Arabic, in general, is a language with a high level of synthesis, that means that it can concentrate within a token more syntactic and grammatical information through the addition of different morphemes.}
In its final version, each token is associated with a total of 11 different annotations, corresponding to the number of the annotation levels we chose.
An excerpt of the corpus after tokens annotation is depicted in table~\ref{tab3}.

For the sake of clarity, in table~\ref{tab3} we show:\\
\textbf{ *} The A column, \textit{Cor}, indicates the token\textipa{"}s source code. For example, the code \textit{3fE}, which stands for \textit{3rab fi Europe}, is the forum from which the text was extracted.\\
\textbf{ *} The B column, \textit{Textco}, is the publication date of the text. \\
\textbf{ *} The C column, \textit{Par}, is the row index of the token in the paragraph.\\
\textbf{ *} The D column, \textit{W}, is the index of the token in the sentence. When \textipa{"}W\textipa{"} corresponds to a range of numbers, it means that the token has been segmented in to its components, specified in the rows below.\\
\textbf{ *} The E column, \textit{Arabi\textipa{S}},  corresponds to the token transcription in Arabish.\\
\textbf{ *} The F column, \textit{Tra}, is the transcription into Arabic characters.\\
\textbf{ *} The G column, \textit{Ita},  is the translation to Italian. \\
\textbf{ *} The H column, \textit{Lem},  corresponds to the lemma.\\
\textbf{ *} The I column, \textit{POS}, is the Part-Of-Speech tag of the token. The tags that have been used for the POS tagging are conform to the annotation system of Universal Dependencies.\\
\textbf{ *} The last three columns (J, K, L) contain the metadata: \textit{Var}, \textit{Age}, \textit{Gen}.


\begin{table}[htbp]\small
\begin{center}
\begin{tabularx}{\columnwidth}{|XlllllX|}
     \hline 
     \textbf{A} & \textbf{B} & \textbf{C} & \textbf{D} & \textbf{E} & \textbf{F} & \textbf{G}\\
     \hline
     \hline
     Cor\label{Cor} & Textco & Par & W & Arabi\textipa{S} & Tra & Ita\\
     \hline
     \hline
     &&&&&&\\
     3fE & 150902 & 2 & 1 & kifech & \RL{كيفاش} & come\\
     
     3fE & 150902 & 2 & 2 & tchou- & \RL{تشوفوا} & vi\\
     & & & & fou & & pare\\
     
     3fE & 150902 & 2 & 3-4 & l3icha & \RL{العيشة} & la vita\\
     
     3fE & 150902 & 2 & 3 & l & \RL{الـ} & -\\
     
     3fE & 150902 & 2 & 4 & 3icha & \RL{عيشة} & -\\
     
     3fE & 150902 & 2 & 5-6 & fil & \RL{فالـ} & all\textipa{"}\\
     
     3fE & 150902 & 2 & 5 & f & \RL{فـ} & -\\
     
     3fE & 150902 & 2 & 6 & il & \RL{الـ} & -\\
     
     3fE & 150902 & 2 & 7 & 4orba & \RL{غربة} & estero\\
     
     3fE & 150902 & 2 & 8 & ? & \RL{؟} & ?\\
     &&&&&&\\
     \hline
     \hline
     & \textbf{H} & \textbf{I} & \textbf{J} & \textbf{K} & \textbf{L} &\\
     \hline
     \hline
     & Lem & POS & Var & Age & Gen &\\
     \hline
     \hline 
     &&&&&&\\
     & \RL{كيفاش} & adv & Bnz & 25-35 & M &\\
     
     & \RL{شاف} & verb & Bnz & 25-35 & M &\\
     
     & \RL{عيشة} & noun & Bnz & 25-35 & M &\\
     
     & \RL {الـ} & det & Bnz & 25-35 & M &\\
     
     & \RL{عيشة} & noun & Bnz & 25-35 & M &\\
     
     & \RL{في} & prep & Bnz & 25-35 & M &\\
     
     & \RL{في} & prep & Bnz & 25-35 & M &\\
     
     & \RL {الـ} & det & Bnz & 25-35 & M &\\
     
     & \RL{غربة} & noun & Bnz & 25-35 & M &\\
     
     & \RL{؟} & pct & Bnz & 25-35 & M &\\
     &&&&&&\\
     \hline 
\end{tabularx}
\caption{An Excerpt of the TArC structure. In the column  \textit{Var}, \textipa{"}Bnz\textipa{"} stands for \textipa{"}Bizerte\textipa{"} a northern city in Tunisia. Glosses: w1:\textit{how}, w2:\textit{do you(pl) see}, w3-4:\textit{the life}, w5-6:\textit{at the}, w7:\textit{outside}, w8:\textit{?}}
\label{tab3}
 \end{center}
\end{table}



Since TA is a spontaneous orthography of TUN, we considered important to adopt the CODA* guidelines as a model to produce a unified lemmatization for each token (column \textit{Lem} in table~\ref{tab3}).
In order to guarantee accurate transcription and lemmatization, we annotated manually the first 6,000 tokens with all the annotation levels.

Some annotation decisions were taken before this step, with regard to specific TUN features:

    \textbf{* Foreign words.} We transcribed the Arabish words into Arabic characters, except for Code-Switching terms. In order to not interrupt the sentences continuity we decide to transcribe Code-Mixing terms into Arabic script. However, at the end of the corpus creation process, these words will be analyzed, making the distinction between acclimatized loans and Code-Mixing. 
    \newline\newline 
    
    The first ones will be transcribed into Arabic characters also in \textit{Lem}, as shown in table~\ref{tab4}. The second ones will be lemmatized in the foreign language, mostly French, as shown in table~\ref{tab5}. 
    \newline\textbf{* Typographical errors.} Concerning typos and typical problems related to the informal writing habits in the web, such as repeated characters to simulate prosodic features of the language, we have not maintained all these characteristics in the transcription (column \textit{Tra}). Logically, these were neither included in \textit{Lem}, according to the CODA* conventions, as shown in table~\ref{tab5}.
    \newline\textbf{* Phono-Lexical exceptions.} We used the grapheme \newline<\RL{ڨ}>, \textipa{[q]}, only in loanword transcription and lemmatization. As can be seen in table~\ref{tab6}, the Hilalian phoneme [g] of the Turkish loanword \textipa{"}gawriyya\textipa{"}, has been transcribed and lemmatized with the grapheme <\RL{ق}>, \textipa{[g]}.
    \newline\textbf{* Glottal stop.} As explained in CODA TUN, real initial and final glottal stops have almost disappeared in TUN. They remain in some words that are treated as exceptions, e.g. <\RL{أسئلة}>, \textipa{["PasPla]}, \textipa{"}question\textipa{"}  \cite{DBLP:conf/lrec/ZribiBMEBH14}. Indeed, we transcribe the glottal stops only when it is usually pronounced, and if it does not, we do not write the glottal stops at the beginning of the word or at the end, neither in the transcription, nor in the lemmas. \\

\begin{table}[htbp]\small
\begin{center}
\begin{tabularx}{\columnwidth}{|XccccX|}
         \hline
         W & Arabi\textipa{S} & Tra & Ita & Lem & POS \\
         \hline
         \hline
         4 & konna & \RL{كنّا} & siamo stati & \RL{كان} & verb \\
         
         5 & far7anin & \RL{فرحانين} & contenti & \RL{فرحان} & adj \\
         
         6 & , & \RL{,} & , & \RL{,} & punct \\
         
         7 & merci & \RL{مرسي} & grazie & \RL{مرسي} & intj \\
         
         \hline

\end{tabularx}
\caption{Loanword example in the corpus. Glosses: w4:\textit{we were}, w5:\textit{happy}, w6:\textit{,} , w7:\textit{thanks}}
\label{tab4}
 \end{center}
\end{table}



\begin{table}[htbp]\small
\begin{center}
\begin{tabularx}{\columnwidth}{|XccccX|} 
         \hline
         W & Arabi\textipa{S} & Tra & Ita & Lem & POS \\
         \hline
         \hline
         1 & R7 & recette & ricetta & recette & noun \\
         
         2 & patee & pâté & patè & pâté & noun \\
         
         3 & dieri & \RL{دياري} & fatto in casa & \RL{دياري} & adj \\
         
         4 & w & \RL{و} & e & \RL{و} & cconj \\
         
         5 & bniiiiin & \RL{بنين} & buonissimo & \RL{بنين} & adj \\
         \hline
         
\end{tabularx}
\caption{Prosody example in the corpus. Glosses: w1:\textit{recipe}, w2:\textit{pâté}, w3:\textit{homemade}, w4:\textit{and}, w5:\textit{delicious}}
\label{tab5}
 \end{center}
\end{table}


\begin{table}[htbp]
\begin{center}
\begin{tabularx}{\columnwidth}{|XccccX|}
         \hline
         W & Arabi\textipa{S} & Tra & Ita & Lem & POS \\ 
         \hline
         \hline
         1 & Mtala9 & \RL{مطلق} & divorziato & \RL{مطلق} & noun \\
         
         2 & min & \RL{من} & da & \RL{من} & noun \\
         
         3 & gawriya & \RL{ڨاورية} & (un\textipa{"})europea & \RL{ڨاوري} & adj \\
         \hline

\end{tabularx}
\caption{Phono-Lexical exceptions in the corpus. Glosses: w1:\textit{divorced}, w2:\textit{from}, w3:\textit{European(f)}}
\label{tab6}
 \end{center}
\end{table}



\textbf{* Negation Marks.} CODA TUN proposes to keep the MSA rule of maintaining a space between the first negation mark and the verb, in order to uniform CODA TUN to the first CODA \cite{habash-etal-2012-conventional}. However, as  \newcite{DBLP:conf/lrec/ZribiBMEBH14} explains, in TUN this rule does not make really sense, but it should be done to preserve the consistency among the various CODA guidelines.  
Indeed, in our transcriptions we report what has been produced in Arabish following CODA TUN rules, while in lemmatization we report the verb lemma. At the same time we segment the negative verb in its minor parts: the circumfix negation marks and the conjugated verb. For the first one, we describe the negative morphological structure in the \textit{Tra} and \textit{Lem} columns, as in table~\ref{tab7}. For the second one, as well as the other verbs, we provide transcription and lemmatization.

\begin{table}[htbp]\small
\begin{center}
\begin{tabularx}{\columnwidth}{|llllll|}
         \hline
         W & Arabi\textipa{S} & Tra & Ita & Lem & POS \\
         \hline
         \hline
         14-15 & manajem- & \RL{ما نجمناش} & non & \RL{نجّم} & verb \\
         & nech & & abbiam &&\\
         & & & potuto &&\\
         
         14 & ma + ch & \RL{ما+ش} & - & \RL{ش}+V+\RL{ما} & part \\
         
         15 & najemne & \RL{نجمنا} & - & \RL{نجّم} & verb \\
         \hline
         
\end{tabularx}
\caption{Circumfix negation marks in the corpus. Glosses: w14-15:\textit{we could not}}
\label{tab7}
 \end{center}
\end{table}


\section{Incremental and Semi-Automatic Transcription}
\label{sec:pro}

In order to make the corpus collection easier and faster, we adopted a semi-automatic procedure based on sequential neural models \cite{DBLP-journals/corr/abs-1904-04733,DinarelliGrobol-Seq2BiseqTransformer-2019}.
Since transcribing Arabish into Arabic is by far the most important information to study the Arabish code-system, the semi-automatic procedure concerns only transcription from Arabish to Arabic script. 

In order to proceed, we used the first group of (roughly) 6,000 manually transcribed tokens as training and test data sets in a 10-fold cross validation setting with 9-1 proportions for training and test, respectively. As we explained in the previous section, French tokens were removed from the data. More precisely, whole sentences containing \emph{non-transcribable} French tokens (code-switching) were removed from the data. 
Since at this level there is no way for predicting when a French word can be transcribed into Arabic and when it has to be left unchanged, French tokens create some noise for an automatic, probabilistic model. After removing sentences with French tokens, the data reduced to roughly 5,000 tokens. We chose this amount of tokens for annotation blocks in our incremental annotation procedure.

We note that by combining sentence, paragraph and token index in the corpus, whole sentences can be reconstructed. However, from 5,000 tokens roughly 300 sentences could be reconstructed, which are far too few to be used for training a neural model.\footnote{Preliminary experiments gave indeed quite poor results, below 50\% token-level accuracy on average.}
Instead, since tokens are transcribed at morpheme level, we split Arabish tokens into characters, and Arabic tokens into morphemes, and we treated each token itself as a sequence.
Our model learns thus to map Arabish characters into Arabic morphemes.

The 10-fold cross validation with this setting gave a token-level accuracy of roughly 71\%. This result is not satisfactory on an absolute scale, however it is more than encouraging taking into account the small size of our data.
This result means that less than 3 tokens, on average, out of 10, must be corrected to increase the size of our corpus. 
With this model we automatically transcribed into Arabic morphemes, roughly, 5,000 additional tokens, corresponding to the second annotation block.
This can be manually annotated in at least 7,5 days, but thanks to the automatic annotation accuracy, it was manually corrected into 3 days.\footnote{We based our estimations of the annotation time needed on the time we spent correcting tokens, which is actually faster because tokens are already transcribed, they don't need to be transcribed from scratch.}
The accuracy of the model on the annotation of the second block was roughly 70\%, which corresponds to the accuracy on the test set.
The manually-corrected additional tokens were added to the training data of our neural model, and a new block was automatically annotated and manually corrected.
Both accuracy on the test set and on the annotation block remained at around 70\%. This is because the block added to the training data was significantly different from the previous and from the third.
Adding the third block to the training data and annotating a fourth block with the new trained model gave in contrast an accuracy of roughly 80\%.
This incremental, semi-automatic transcription procedure is in progress for the remaining blocks, but it is clear that it will make the corpus annotation increasingly easier and faster as the amount of training data will grow up.

Our goal concerning transcription, is to have the 25,000 tokens mentioned in section~\ref{subsec:txt-coll} annotated automatically and manually corrected. These data will constitute our gold annotated data, and they will be used to automatically transcribe further data.

\section{Conclusions}
\label{sec:concl}

In this paper we presented TArC, the first Tunisian Arabish Corpus annotated with morpho-syntactic information. We discussed the decisions taken in order to highlight the phonological and morphological features of TUN through the TA corpus structure. Concerning the building process, we have shown the steps undertaken and our effort intended to make the corpus as representative as possible of TA. We therefore described the texts collection stage, as well as the corpus building and the semi-automatic procedure adopted for transcribing TA into Arabic script, taking into account CODA* and CODA TUN guidelines. At the present stage of research, TArC consists of 25.000 tokens, however our work is in progress and for future research we plan to enforce the semi-automatic transcription, which has already shown encouraging results (accuracy = 70\%). We also intend to realize a semi-automatic TA Part-Of-Speech tagger. 
Thus, we aim to develop tools for TA processing and, in so doing, we strive to complete the annotation levels (transcription, POS tag, lemmatization) semi-automatically in order to increase the size of the corpus, making it available for linguistic analyses on TA and TUN.





\section{Bibliographical References}
\label{reference}
\bibliographystyle{lrec2020}
\bibliography{lrec2020W-xample-kc}

\section{Language Resource References}
\bibliographylanguageresource{lrec2020W-xample-kc}

\end{document}